\def\year{2020}\relax
\documentclass[letterpaper]{article} 
\usepackage{aaai20}  
\usepackage{times}  
\usepackage{helvet} 
\usepackage{courier}  
\usepackage[hyphens]{url}  
\usepackage{graphicx} 
\usepackage{graphicx}  
\usepackage{latexsym}
\usepackage{comment}
\usepackage{amsmath}
\usepackage{amsfonts}
\usepackage{booktabs}
\usepackage{xcolor}

\usepackage{url}

\newcommand{\tabref}[1]{Table \ref{#1}}
\newcommand{\figref}[1]{Figure \ref{#1}}
\renewcommand{\eqref}[1]{Eq.~\ref{#1}}

\DeclareMathOperator{\rel}{\boldsymbol{rel}}

\DeclareMathOperator{\atten}{atten}
\DeclareMathOperator{\softmax}{softmax}

\DeclareMathOperator{\A}{\boldsymbol{Agr}}
\DeclareMathOperator{\D}{\boldsymbol{Dis}}
\DeclareMathOperator{\N}{\boldsymbol{N}}

\title{Automatic Fact-Guided Sentence Modification}

\author{Darsh J Shah,* ~~ Tal Schuster,*  ~~ Regina Barzilay\\
  \mbox{}
Computer Science and Artificial Intelligence Lab\\
Massachusetts Institute of Technology \\
\{darsh, tals, regina\}@csail.mit.edu
}  
 
\begin{document}

\maketitle
\let\svthefootnote\thefootnote
\let\thefootnote\relax\footnote{\textbf{*} Order decided by a coin toss.}
\addtocounter{footnote}{-1}
\let\thefootnote\svthefootnote
\begin{abstract}
Online encyclopediae like Wikipedia contain large amounts of text that need frequent corrections and updates. The new information may contradict existing content in encyclopediae. In this paper, we focus on rewriting such dynamically changing articles. This is a challenging constrained generation task, as the output must be consistent with the new information and fit into the rest of the existing document. To this end, we propose a two-step solution: (1) We identify and remove the contradicting components in a target text for a given claim, using a neutralizing stance model; (2) We expand the remaining text to be consistent with the given claim, using a novel two-encoder sequence-to-sequence model with copy attention. Applied to a Wikipedia fact update dataset, our method successfully generates updated sentences for new claims, achieving the highest SARI score. Furthermore, we demonstrate that generating synthetic data through such rewritten sentences can successfully augment the FEVER fact-checking training dataset, leading to a relative error reduction of 13\%.\footnote{ Code: (1) \url{https://github.com/TalSchuster/TokenMasker}\\ (2) \url{https://github.com/darsh10/split_encoder_pointer_summarizer}}

\end{abstract}
\section{Introduction}
\label{sec:introduction}
Online text resources like Wikipedia contain millions of articles that must be continually updated. Some updates involve expansions of existing articles, while others modify the content. In this work, we are interested in the latter scenario where the modification contradicts the current articles. Such changes are common in online sources and often cover a broad spectrum of subjects ranging from the changing of dates for events to modifications of the relationship between entities. In these cases, simple solutions like negating the original text or concatenating it with the new information would not apply. In this work, our goal is to automate these updates. Specifically, given a claim and an outdated sentence from an article, we rewrite the sentence to be consistent with the given claim while preserving non-contradicting content.

\begin{figure}[t]
  \centering
\includegraphics[width=0.47\textwidth]{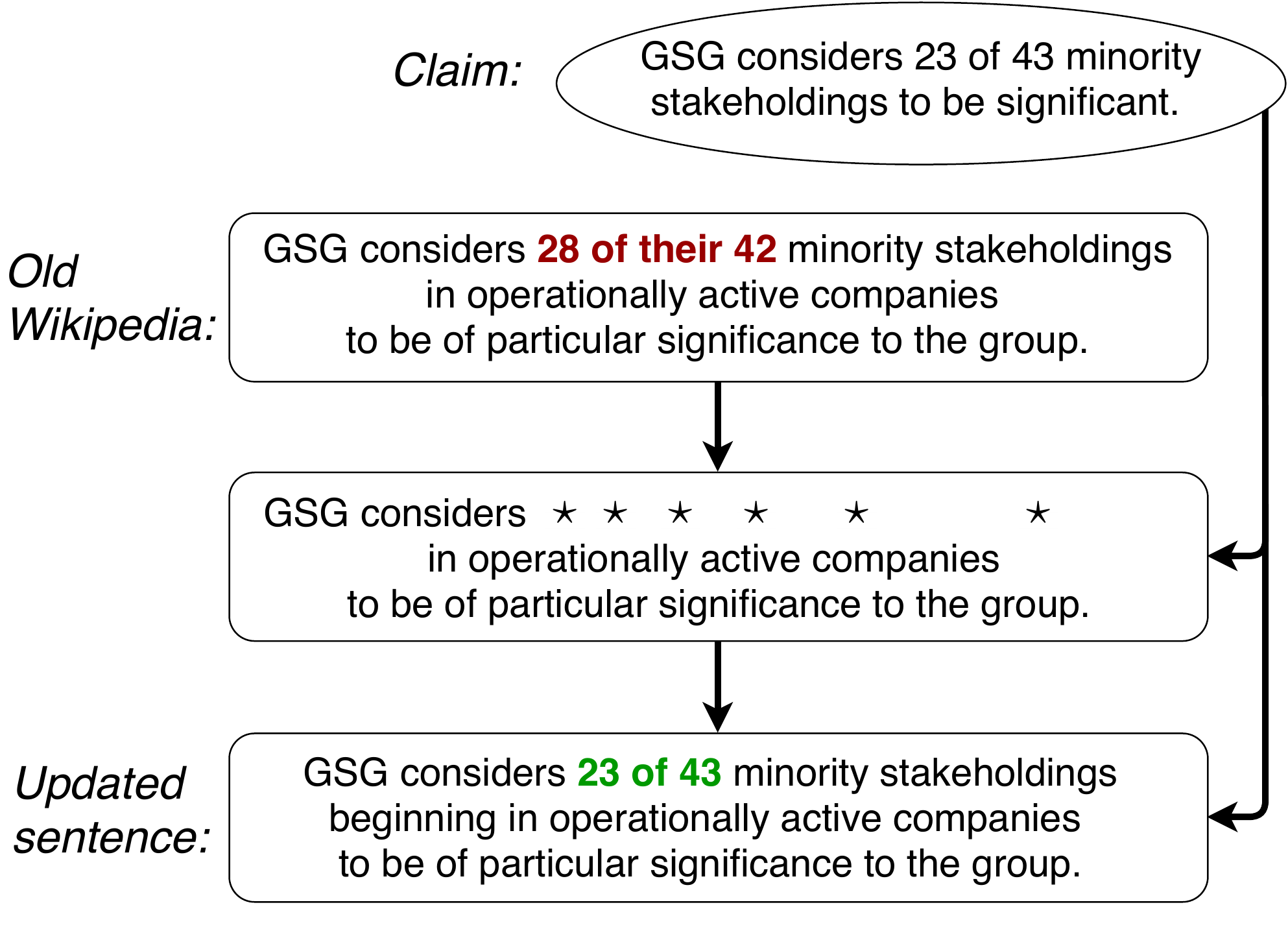}
\caption{Our fact-guided update pipeline. Given a claim which refutes incorrect information, a masker is applied to remove the contradicting parts from the original text while preserving the rest of the context. Then, the residual neutral text and claim are fused to create an updated text that is consistent with the claim.
}\label{fig:Obama_example}
\end{figure}

Consider the Wikipedia update scenario depicted in Figure \ref{fig:Obama_example}. The claim, informing that \textit{23 of 43} minority stakeholdings are significant, contradicts the old information in the Wikipedia sentence, requiring modification. Directly learning a model for this task would demand supervision, i.e.\ demonstrated updates with the corresponding claims. For Wikipedia, however, the underlying claims which drive the changes are not easily accessible. Therefore, we need to utilize other available sources of supervision. 

In order to make the corresponding update, we develop a two step solution: (1) Identify and remove the contradicting segments of the text (in this case, \textit{28 of their 42 minority stakeholdings}); (2) Rewrite the residual sentence to include the updated information (e.g.\ fraction of significant stakeholdings) while also preserving the rest of the content. 

For the first step, we utilize a neutrality stance classifier as indirect supervision to identify the polarizing spans in the target sentence. We consider a sentence span as polarizing if its absence increases the neutrality of the claim-sentence pair. To identify and mask such sentence spans, we introduce an interpretability-inspired ~\cite{lei-rational} neural architecture to effectively explore the space of possible spans. We formulate our objective in a way that the masking is minimal, thus preserving the context of the sentence. 

For the second step, we introduce a novel, two-encoder decoder architecture, where two encoders fuse the claim and the residual sentence with a more refined control over their interaction.

We apply our method to two tasks: automatic fact-guided modifications and data augmentation for fact-checking. On the first task, our method is able to generate corrected Wikipedia sentences guided by unstructured textual claims. Evaluation on Wikipedia modifications demonstrates that our model's outputs were the most successful in making the requisite updates, compared to strong baselines. On the FEVER fact-checking dataset, our model is able to successfully generate new claim-evidence supporting pairs, starting with claim-evidence refuting pairs --- intended to reduce the bias in the dataset. Using these outputs to augment the dataset, we attain a 13\% decrease in relative error on an unbiased evaluation set.
\section{Related Work}
\label{sec:related_work}
\paragraph{Text Rewriting}

There have been several recent advancements in the field of text rewriting, including style transfer
~\cite{shen2017style,zhang2018style,chen-etal-2018-learning} and sentence fusion ~\cite{barzilay2005sentence,narayan-etal-2017-split,geva2019discofuse}. Unlike previous approaches, our sentence modification task addresses potential contradictions between two sources of information.

Our work is fairly related to the approach of \cite{li-etal-2018-delete}, which separates the task of sentiment transfer into deleting strong markers of sentiment in a sentence and retrieving markers of the target label to generate a sentence with the opposite sentiment. In contrast to such work, where the requisite modification is along a fixed aspect (e.g.\ sentiment), in our setting, an arbitrary input sentence (the claim) dictates the space of desired modifications. Therefore, in order to succeed at our task, a system should understand the varying degree of polarization in the spans of the outdated sentence against the claim before modifying the sentence to be consistent with the claim.

\paragraph{Wikipedia Edits}
Wikipedia edit history has been analyzed for insights into the kinds of modifications made
\cite{daxenberger-gurevych-2013-automatically,yang-etal-2017-identifying-semantic,faruqui-etal-2018-wikiatomicedits}. The edit history has also been used for text generation tasks such as sentence compression and simplification
\cite{yatskar-etal-2010-sake}, paraphrasing \cite{max-wisniewski-2010-mining} and writing assistance 
\cite{cahill-etal-2013-robust}. In this work, we are interested in the novel task of automating the editing process with the guidance of a textual claim.

\paragraph{Fact Verification Datasets}

The growing interest in automatic fake news detection led to the development of several fact verification datasets 
~\cite{vlachos-riedel-2014-fact,wang-2017-liar,rashkin-etal-2017-truth,fever}. FEVER, the largest fact-checking dataset, contains 185K human written fake and real claims, generated by crowd-workers, in context of sentences from Wikipedia articles. 
This dataset contains biases that allow a model to identify many of the false claims without any evidence~\cite{schuster2019towards}. 
This bias affects the generalization capabilities of models trained on such data. 
In this work, we show that our automatic modification method can be used to augment a fact-checking dataset and to improve the inference of models trained on it.

\paragraph{Data Augmentation}
Methods for data augmentation are commonly used in computer vision~\cite{perez2017effectiveness}. There have been recent successes in NLP where augmentation techniques such as paraphrasing and word replacement were applied to text classification ~\cite{kobayashi2018contextual,wu2018conditional}. Adversarial examples in NLI with syntactic modifications can also be considered as methods of data augmentation ~\cite{iyyer-etal-2018-adversarial,zhang2019paws}.
 In this work, we create constrained modifications, based on a reference claim, to augment data for our task at hand. Our additions are specifically aimed towards reducing the bias in the training data, by having a false claim appear in both ``Agrees'' and ``Disagrees'' classes.

\section{Model}
\paragraph{Problem Statement}
\label{sec:prob_state}

We assume access to a corpus $\mathcal{D}$ of claims and knowledge-book sentences. Specifically, $\mathcal{D} = \{\{C_1 ,..., C_n\}, \{S_1, ..., S_m\}\}$, where $C$ is a short factual sentence (claim), and $S$ is a sentence from Wikipedia. Each pair of claim and Wikipedia sentence has a relation $\rel(S, C)$, of either agree ($\A$), disagree ($\D$) or neutral ($\N$). In this corpus, a Wikipedia sentence $S$ is defined as outdated with respect to $C$ if $\rel(S,C)=\D$ and updated if $\rel(S,C)=\A$. The neutral relation holds for pairs in which the sentence doesn't contain specific information about the claim.

Our goal is to automatically update a given sentence $S$, which is outdated with respect to a $C$. Specifically, given a claim and a pair for which $\rel(S, C) = \D$, our objective is to apply minimal modifications to $S$ such that the relation of the modified sentence $S^+$ will be: $\rel(S^+, C) = \A$. In addition, $S^+$ should be structurally similar to $S$.

\paragraph{Framework}

\begin{figure*}[!t]
\centering
\includegraphics[width=0.95\textwidth]{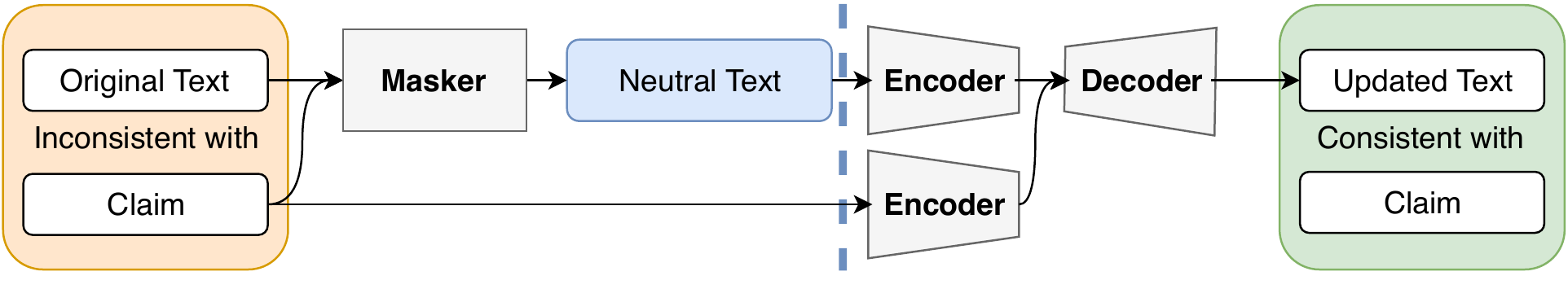}
\caption{A summary of our pipeline. Given a sentence that is inconsistent with a claim, a masker is applied to mask out the contradicting parts from the original text while preserving the rest of the content. Then, the residual neutral text and claim are fused to create an updated text that is consistent with the claim. The Masker and the Two-Encoder Generator are trained separately.
 \label{fig:pipeline}}
\end{figure*}

Currently, to the best of our knowledge, there is no large dataset for fact-guided modifications. Instead, we utilize a large dataset with pairs of claims and sentences that are labeled to be consistent, inconsistent or neutral.
In order to compensate the lack of direct supervision, we develop a two-step solution. First, using a pretrained fact-checking classifier for indirect supervision, we identify the polarizing spans of the outdated sentence and mask them to get a $S^{\emptyset}$ such that $\rel(S^{\emptyset},C) = \N$. Then, we fuse this pair to generate the updated sentence which is consistent with the claim. This is done with a sequence-to-sequence model trained with consistent pairs through an auto-encoder style objective. The two steps are trained independently to simplify optimization (see \figref{fig:pipeline}).

\begin{figure*}[!t]
\centering
\includegraphics[width=0.95\textwidth]{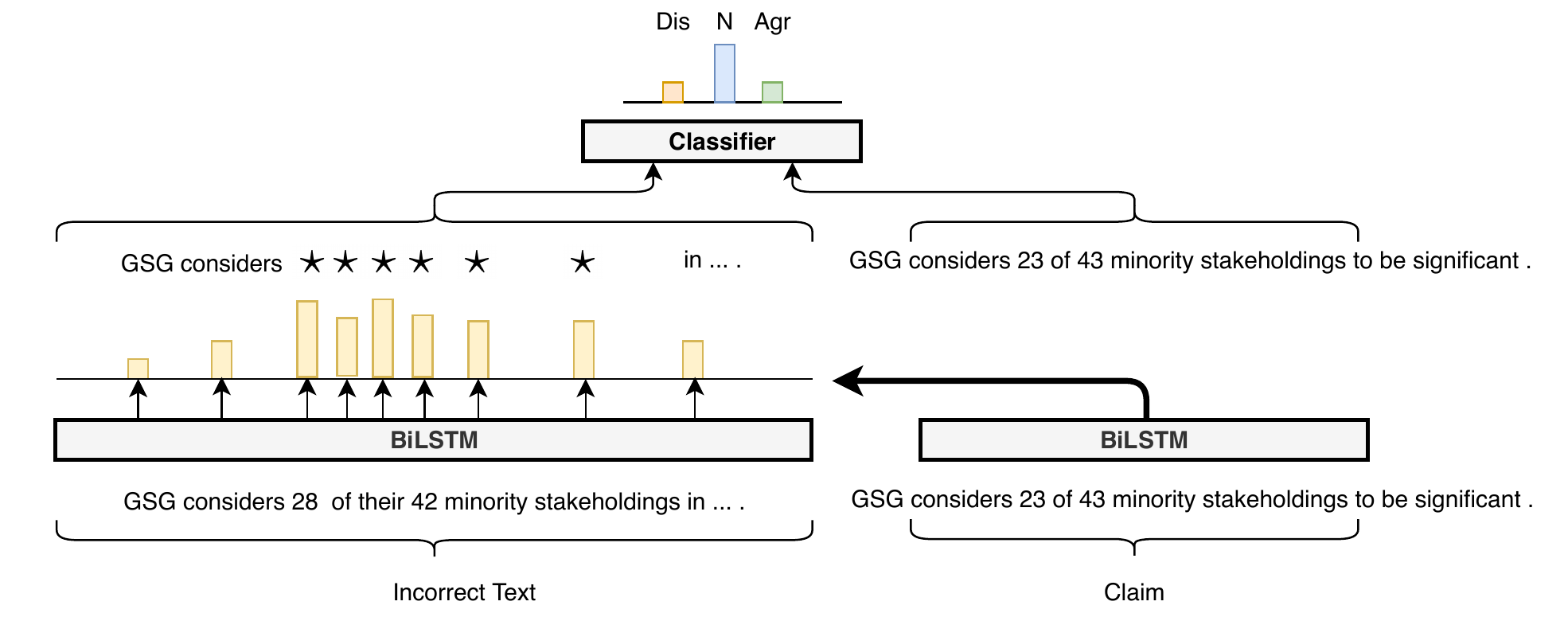}
\caption{Illustrating the flow of the masker module.
}\label{fig:mask_gen}
\end{figure*}
\subsection{Masker: Eliminate Polarizing Spans} \label{sec:mask_gen}

In this section we describe the module to identify the polarizing spans within a Wikipedia sentence. Masking these spans ensures that the residual sentence-claim pairs attain a neutral relation. Here, neutrality is determined by a classifier trained on claim and Wikipedia sentence pairs as described below. Using this classifier, the masking module is trained to identify the polarizing spans by maximizing the neutrality of the residual-sentence and claim pairs. In order to preserve the context of the original sentence, we include optimization constraints to ensure minimal deletions. This approach is similar to neural rationale-based models~\cite{lei-rational}, where a module tries to identify the spans of the input that justify the model's prediction.

\paragraph{Neutrality Masker} Given a knowledge-book sentence ($S$) and a claim ($C$), the masker's goal is to create $S^{\emptyset}$ such that $\rel(S^{\emptyset},C) = \N$. For the original sentence with $l$ tokens, $S = \{x_i\}_{i=1}^{l}$, the output is a mask $m\in[0,1]^l$. The neutral sentence $S^{\emptyset}$ is constructed as:
\begin{equation}
  {S^\emptyset_i}=\begin{cases}
    x_i, & \text{if $m_i=0$}\\
    \star, & \text{otherwise}
  \end{cases}
\end{equation}
where $\star$ is a special token.\footnote{The special token is treated as an out-of-vocabulary token for the following models.} The details of the masker architecture are stated below and depicted in \figref{fig:mask_gen}.

\paragraph{Encoding}
We encode $S$ with a sequence encoder to get $e_i = f(x; \boldsymbol{w}_f)_i$.
Since the neutrality of the sentence needs to be measured with respect to a claim, we also encode the claim and enhance $S$'s representations with that of $C$ using attention mechanism. Formally, we compute
\begin{equation}
	z_i = e_i + \sum_{j=1}^n a_{i,j}\cdot c_j , 	
\end{equation}
where $c_j$ are the encoded representations of the claim and $a_{i,j}$ are the parameterized bilinear attention ~\cite{kim2018bilinear} weights computed by:
\begin{equation}
	a_{i,j} = \softmax_j(\atten(e_i, c_j)),
\end{equation}

\begin{equation}
	\atten(e_i, c_j) = e_i W c_j^T + b.
\end{equation}
Finally, the aggregated representations are used as input to a sequence encoder $g(\cdot; \boldsymbol{w}_g)$.

\paragraph{Masking}

The encoded sentence is used to predict a per token masking probability:
\begin{equation}
	p(m_i = 1) = \sigma(g(z; \boldsymbol{w}_g)_i).
\label{eq:mask_prob}
\end{equation}

Then, the mask is applied to achieve the residual sentence:
\begin{equation}
    S^{\emptyset} = S \circ (1-m),
\end{equation}
where $\circ$ denotes element-wise multiplication. During training, we perform soft deletions over the token embeddings and add the out-of-vocabulary embedding in place. During inference, the values of $m$ are rounded to create a discrete mask.

\paragraph{Training}
A pretrained fact-checking neutrality classifier's prediction $\rel(S,C)$ is used to guide the training of the masker. In order to encourage maximal retention of the context, we utilize a regularization term to minimize the fraction of the masked words. The joint objective is to minimize:
\begin{equation}
    \small{
    \mathcal{L}(S,C,m)\!= - \log \left(p(\rel(S^\emptyset,C){=}\N)\right) + \frac{\lambda}{l} \sum_{i=1}^{l} m_i.
    }
\label{eq:mask_gen_loss} 
\end{equation}

\paragraph{Fact-checking Neutrality Classifier} \label{sec:neutral_classifier}
 Our fact-checking classifier is pretrained on agreeing and disagreeing $(S,C)$ pairs from $\mathcal{D}$, in addition to neutral examples constructed through negative sampling. For each claim we construct a \textit{neutral} pair by sampling a random sentence from the same paragraph of the polarizing sentence, making it contextually close to the claim, but unlikely to polarize it. We pretrain the classifier on these examples and fix its parameters during the training of the masker.

\paragraph{Optional Syntactic Regularization}

Currently the model is trained with distant supervision, so, we pre-compute a valid neutrality mask as additional signal, when possible. To this end, we parse the original sentences using a constituency parser and iterate over continuous syntactic phrases by increasing length. For each sentence, the shortest successful neutrality mask (if any) is selected as a target mask.\footnote{If there are several successful masks of the same length, we use the one with the highest neutrality score.} In the event of successfully finding such a mask, the masking module is regularized to emulate the target mask by adding the following term to \eqref{eq:mask_gen_loss}:
\begin{equation}
    \frac{1}{l} || m - m' ||^2,
    \label{eq:boots}
\end{equation}
where $m'$ is the target mask.

Empirically, we find that the model can perform well even without this regularization, but it can help to stabilize the training. Additional details and analysis are available in the appendix.

\subsection{Two-encoder Pointer Generator: Constructing a Fact-updated Sentence} \label{sec:pointer_gen}
In this section we describe our method to generate an output which agrees with the claim. If the earlier masking step is done perfectly, the merging boils down to a simple fusion task. However, in certain cases, especially ones with a strong contradiction, our minimal deletion constraint might leave us with some residual contradictions in $S^{\emptyset}$. Thus, we develop a model which can control the amount of information to consider from either input.

We extend the pointer-generator model of \cite{pointer-generator} to enable multiple encoders. While sequence-to-sequence models support the encoding of multiple sentences by simply concatenating them, our use of a per input encoder allows the decoder to better control the use of each source. This is especially of interest to our task, where the context of the claim must be translated to the output while ignoring contradicting spans from the outdated Wikipedia sentence. 

Next, we describe the details of our generator's architecture. Here, we use one encoder for the outdated sentence and one encoder for the claim. In order to reduce the size of the model, we share the parameters of the two encoders. The model can be similarly extended to any number of encoders.


\paragraph{Encoding}
At each time step $t$, the decoder output $h^t$, is a function of a weighted combination of the two encoders' context representations $r^{t}$, the decoder output in the previous step $h^{t-1}$ and the representation of the word output at the end of the previous step $emb(y^{t-1})$: 

\begin{equation}
    h^t = \textit{RNN}([r^t, emb(y^{t-1})], h^{t-1}).
    \label{eq:decoder}
\end{equation}

As the decoder should decide at each time step which encoder to attend more, we introduce an encoder weight $\alpha$. The shared encoder context representation $r^t$ is based on their individual representations $r^t_1$ and $r^t_2$: 

\begin{align*}
    \alpha = \sigma(u^{T}_{enc}[r^t_{1}, r^t_{2}]), \\
    r^t = \alpha \cdot r^t_{1} + (1-\alpha)r^t_{2}.
    \tag{10}
\end{align*}

The context representation $r_i^t$ ($i{\in}\{1,2\}$) is the attention score over the encoder representation $r_i$ for a particular decoder state $h^{t-1}$:

\begin{align*}
    z_j^{t} &= u^{T}\tanh(r_{i,j} + h^{t-1}), \\
    a_{i}^t &= \softmax(z^t), \\
    r_i^t &= \sum_{j}a_{i,j}^{t}{r_{i,j}}. 
    \tag{11} \label{eq:rep}
\end{align*}

\paragraph{Decoding}
Following standard copy mechanism, predicting the next word $y^t$, involves deciding whether to \textit{generate} ($p_{gen}$) or \textit{copy}, based on the decoder input $x^{t} = [r^t, emb(y^{t-1})]$, the decoder state $h^{t}$ and context vector $r^{t}$:

\begin{equation*}
    p_{gen} = \sigma(v^{T}_{x}x^{t} + v^{T}_{h}h^{t} + v^{T}_{r}r^{t}).
    \label{eq:generate_copy}
    \tag{12} 
\end{equation*}

In case of copying, we need an additional gating mechanism to select between the two sources: 

\begin{equation*}
    p_{enc1} = \sigma(u^{T}_{x}x^{t} + u^{T}_{h}h^{t} + u^{T}_{r}r^{t}).
    \label{eq:encoder_voc}
    \tag{13} 
\end{equation*}

When generating a new word, the probability over words from the vocabulary is computed by:

\begin{equation*}
    P_{vocab} = \softmax(V^{T}[h^t,r^t]).
    \label{eq:vocab}
    \tag{14} 
\end{equation*}

The final output of the decoder at each time step is then computed by:

\begin{align*}
    P(w) &= p_{gen}P_{vocab}(w)  + \\  
    & (1-p_{gen})(p_{enc1})\sum_{j:w_{j}=w} a_{1,j}^{t} + \\
    & (1-p_{gen})(1 - p_{enc1})\sum_{j:w_{j}=w} a_{2,j}^{t}, \\ 
     y^t = & \operatorname{argmax}_w P(w).
     \tag{15} \label{eq:pointer_out}
\end{align*}   
where $a^t$ are the input sequence attention scores from \eqref{eq:rep}.

\paragraph{Training} 

Since we have no training data for claim guided sentence updates, we train the generator module to reconstruct a sentence $S$ to be consistent with an agreeing claim $C$. The training input is the residual up-to-date neutral sentence $S^{\emptyset}$ and the guiding claim $C$. 

During inference, we utilize only guiding claims and residual outdated sentences $S^{\emptyset}$ to create $S^+$. While generating the updated sentences $S^+$, we would like to preserve as much context as possible from the contradicting sentence, while ensuring the correct relation with the claim. Therefore, for each case, if the later goal is not achieved, we gradually increase the focus on the claim by increasing $\alpha$ and $p_{enc1}$ values until the output $S^+$ satisfies $\rel(S^+,C)=\A$, or until a predefined maximum weight.



\section{Experimental Setup}
\label{sec:experiments}
We evaluate our model on two tasks: (1) Automatic fact updates of Wikipedia sentences, where we update outdated wikipedia sentences using guiding fact claims; and (2) Generation of synthetic claim-evidence pairs to augment an existing biased fact-checking dataset in order to improve the performance of trained classifiers on an unbiased dataset.

\subsection{Datasets}

\paragraph{Training Data from FEVER}
We use FEVER \cite{fever}, the largest available Wikipedia based fact-checking dataset to train our models for both of our tasks. This dataset contains claim-evidence pairs where the claim is a short factual sentence and the evidence is a relevant sentence retrieved from Wikipedia.
We use these pairs as our claim-setnence samples and use the ``refutes'', ``not enough information'', ``supports'' labels of that dataset as our $\D, \N, \A$ relations, respectively.

\paragraph{Evaluation Data for Automatic Fact Updates}
We evaluate the automatic fact updates task on an evaluation set based on part of the symmetric dataset from ~\cite{schuster2019towards} and the fact-based cases from a Wikipedia updates dataset \cite{yang-etal-2017-identifying-semantic}. For the symmetric dataset, we use the modified Wikipedia sentences with their guiding claims to generate the true Wikipedia sentence. For the cases from the updates dataset, we have human annotators write a guiding claim for each update and use it, together with the outdated sentence, to generate the updated Wikipedia sentence.
Overall we have a total of 201 tuples of fact update claims, outdated sentences and updated sentences.

\paragraph{Evaluation Data for Augmentation}

To measure the proficiency of our generated outputs for data augmentation, we use the unbiased FEVER-based evaluation set of \cite{schuster2019towards}.
As shown by \cite{schuster2019towards}, the claims in the FEVER dataset contain give-away phrases that can make FEVER-trained models overly rely on them, resulting in decreased performance when evaluated on unbiased datasets. 

The classifiers trained on our augmented dataset are evaluated on the unbiased symmetric dataset of \cite{schuster2019towards}. This dataset (version 0.2) contains 531 claim-evidence pairs for validation and 534 claim-evidence pairs for testing.

In addition, we extend the symmetric test set by creating additional FEVER-based pairs.
We hired crowd-workers on Amazon Mechanical Turk and asked them to simulate the process of generating synthetic training pairs. Specifically, for a ``refutes'' claim-evidence FEVER pair, the workers were asked to generate a modified supporting evidence while preserving as much information as possible from the original evidence. We collected responses of workers for 500 refuting pairs from the FEVER training set. 
This process extends the symmetric test set (\textsc{+TURK}) by 1000 cases --- 500 ``refutes'' pairs, and corresponding 500 ``supports'' pairs generated by turkers.

\begin{table*}[t]
\small
\centering
\begin{tabular}{llrrrrccc}
\toprule
& & \multicolumn{4}{c}{Automatic Evaluation} & & \multicolumn{2}{c}{Human's Scores}\\
\cmidrule(lr){3-6}\cmidrule(lr){8-9}
&\textsc{model}              & \textsc{SARI} & \textsc{Keep} & \textsc{Add} & \textsc{Del} & & \textsc{Grammar} & \textsc{Agreement} \\ 
\midrule
\multicolumn{2}{l}{\textit{\textbf{Fact updates}}:} &\\
&Split-no-Copy & 15.1 & 36.9 & 1.9 & 49.5 && - & -\\
&Paraphrase & 15.9 & 18.7 & 4.2 & 50.7 & & 3.75 & 3.65 \\
&Claim Ext.    & 12.9 & 22.6       & 1.9      & 50.4 & & 1.75 & 2.65     \\ 
&M. Concat     & 26.5 & \textbf{61.7}      & 6.7      & 44.9  & & 3.28 & 2.75 \\ 
&Ours & \textbf{31.5} & 45.4      & \textbf{13.2}      & \textbf{52.1}  & & \textbf{3.85} & \textbf{4.00} \\ 
&Human &\multicolumn{4}{r}{\rule[0.09cm]{4cm}{0.01cm}} & & 4.80 & 4.70 \\
\midrule

\multicolumn{2}{l}{\textit{\textbf{Data augmentation}}:} &\\
&Paraphrase & 18.2 & 12.5 & 10.6 & 45.7 & & 4.12 & 3.92 \\
&Claim Ext.    & 12.2 & 9.8       & 4.0      & 46.4  & & 1.58 & 2.84    \\ 
&M. Concat     & 22.1 & \textbf{71.6}      & 6.8      & 22.3  & & \textbf{4.45} & 2.05  \\ 
&Ours               & \textbf{34.4} & 33.0      & \textbf{26.0}     & \textbf{47.5}   & & 4.14 & \textbf{3.98}  \\
&Human &\multicolumn{4}{r}{\rule[0.09cm]{4cm}{0.01cm}} & & 4.69 & 4.15 \\

\bottomrule
\end{tabular}
\caption{Human evaluation results for our model's outputs for the fact update task (top) and for the data augmentation task (bottom). 
The left part of the table shows the geometric SARI score with the three \textsc{F1} scores that construct it. 
The right part shows the human's scores in a 1-5 Likert scale on grammatically of the output sentence and on agreement with the given claim.}\label{tab:symmetric_auto_res}

\end{table*}

\subsection{Implementation Details}

\paragraph{Masker}
We implemented the masker using the AllenNLP framework \cite{Gardner2017AllenNLP}. For a neutrality classifier, we train an ESIM model \cite{chen2017enhanced} to classify a relation of $\A$, $\D$ or $\N$. To train this classifier, we use the $\A$ and $\D$ pairs from the FEVER dataset and for each claim we add a neutral sentence which is sampled from the sentences in the same document as the polarizing one. The classifier and masker are trained with GloVe~\cite{pennington-etal-2014-glove} word embeddings. We use BiLSTM \cite{sak2014long} encoders with hidden dimensions of 100 and share the parameters of the claim and original sentence encoders. The model is trained for up to 100 epochs with a patience value of 10, where the stopping condition is defined as the highest delta between accuracy and deletion size on the development set ($\Delta$ in \tabref{tab:mask_res}).

For syntactic guidance, we use the constituency parser of \cite{stern-etal-2017-minimal} and consider continuous spans of length 2 to 10 as masking candidates (without combinations). By doing so, we obtain valid neutrality masks for 38\% of the $\A$ and $\D$ pairs from the FEVER training dataset. These masks are used for \eqref{eq:boots}.

\paragraph{Two-Encoder Pointer Generator}
We implemented our proposed multi-sequence-to-sequence model, based on the pointer-generator framework.
We use a one layer BiLSTM for encoding and decoding with a hidden dimension of 256. The parameters of the two encoders are shared. The model is trained with batches of size 64 for a total of 50K steps.

\paragraph{BERT Fact-Checking Classifier}
We use a BERT \cite{devlin2018bert} classifier, which takes in as input a (claim-evidence) pair separated by a special token, to predict out of 3 labels ($\A$, $\D$ or $\N$). The model is fine-tuned for 3 epochs, which is sufficient to perform well on the task.

\paragraph{Evidence Regeneration}
Since we are interested in using the generated supporting pairs for data augmentation, we add machine generated cases to the $\A$ set of the dataset. Adding machine generated sentences to only one of the labels in the data can be ineffective. Therefore, we balance this by regenerating paraphrased refuting evidence for the false claims. This is then added along with all models' outputs for a balanced augmentation.

\subsection{Baselines}
We consider the following baselines for constructing a fact-guided updated sentence:

\begin{itemize}
\item \textbf{Copy Claim} The sentence of the claim is copied and used as the updated sentence for itself (used only for data augmentation).

\item \textbf{Paraphrase} The claim is paraphrased using the back-translation method of \cite{wieting-gimpel-2018-paranmt}\footnote{\url{https://github.com/vsuthichai/paraphraser}}, and the output is used as the updated sentence.

\item \textbf{Claim Extension [Claim Ext.]} A pointer-generator network is trained to generate the updated sentence from an input claim alone. The model is trained on FEVER's agreeing pairs and applied on the to-be-updated claims during inference.

\item \textbf{Masked Concatenation [M. Concat]} Instead of our Two-Encoder Generator, we use a pointer-generator network.
The residual sentence (output from the masker module) and the claim are concatenated and used as input.

\item \textbf{Split Encoder without Copy [Split-no-Copy]} Our Two-Encoder Generator, without the copy mechanism. The original text and contradicting claim are passed through each of the encoders.

\end{itemize}
\section{Results}
\label{sec:results}
We report the performance of the model outputs for automatic fact-updates by comparing them to the corresponding correct wikipedia sentences. We also have crowd workers score the outputs on grammar and for agreeing with the claim. Additionally, we report the results on a fact-checking classifier using model outputs from the FEVER training set as data augmentation.

\begin{table}[t]
\centering
\begin{tabular}{lccc}
\toprule
\textsc{Model} & \textsc{Dev} & \textsc{Test}   & \textsc{+Turk}\\ 
\midrule
No Augmentation & 62.7 & 66.1 & 77.0\\
\midrule
Paraphrase & 60.8 & 64.6 & 77.4 \\
Copy Claim    &     62.1     &  63.6 & 77.4    \\
Claim Ext.  &     62.5              &  65.0 & 76.8\\
M. Concat     &    60.1   & 63.7  & 78.5\\ 
\midrule
Ours & \textbf{63.8} & \textbf{67.8} & \textbf{80.0} \\
\bottomrule
\end{tabular}
\caption{Classifiers' accuracy on the symmetric \textsc{Dev} and \textsc{Test} splits. The right column (\textsc{+Turk}) shows the accuracy on the \textsc{Test} set extended to include the 500 responses of turkers for the simulated process and the refuted pairs that they originated from. The BERT classifiers were trained on the FEVER training dataset augmented by outputs of the different methods.}
\label{tab:data_augmentation3}
\end{table}

\paragraph{Fact Updates} 
Following recent text simplification work, we use the SARI~\cite{xu-etal-2016-optimizing} method. The SARI method takes 3 inputs: (i) original sentence, (ii) human written updated sentence and (iii) model output. It measures the similarity of the machine generated and human reference sentences based on the deletions, additions and kept n-grams\footnote{We use the default up to 4-grams setting.} with respect to the original sentence.\footnote{Following \cite{geva2019discofuse} we use the F1 measure for all three sets, including deletions. The final \textsc{SARI} score is the geometric mean of the \textsc{Add}, \textsc{Del} and \textsc{Keep} score.} 
For human evaluation of the model's outputs, 20\% of the evaluation dataset was used.
Crowd-workers were provided with the model outputs and the corresponding supposably consistent claims. They were instructed to score the model outputs from 1 to 5 (1 being the poorest and 5 the highest), on grammaticality and agreement with the claim.

Table \ref{tab:symmetric_auto_res} reports the automatic and human evaluation results. Our model gets the highest \textsc{SARI} score, showing that it is the closest to humans in modifying the text for the corresponding tasks. Humans also score our outputs the highest for consistency with the claim, an essential criterion of our task. In addition, the outputs are more grammaticality sound compared to those from other methods. 

Examining the gold answers, we notice that many of them include very minimal and local modifications, keeping much of the original sentence.  The M.\ Concat model keeps most of the original sentence as is, even at the cost of being inconsistent with the claim. This corresponds to a high \textsc{Keep} score but a lower \textsc{SARI} score overall, and a low human score on supporting the claim. Claim Ext.\ and Paraphrase do not maintain the structure of the original sentence, and perform poorly on \textsc{Keep}, leading to a low \textsc{SARI} score. The Split-no-Copy model has the same low \textsc{ADD} score as Claim Ext.\ since instead of copying the accurate information from the claim, it generates other tokens.

\paragraph{Data Augmentation} \label{res:aug}
For 41850 $\D$ pairs in the FEVER training data, our method generates synthetic evidence sentences leading to 41850 $\A$ pairs. We train the BERT fact-checking classifier with this augmented data and report the performance on the symmetric dataset in Table \ref{tab:data_augmentation3}. In addition, we repeat the human evaluation process on the generated augmentation pairs and report it in Table \ref{tab:symmetric_auto_res}.

Our method's outputs are effective for augmentation, outperforming a classifier trained only on the original biased training data by an absolute 1.7\% on the \textsc{Test} set and an absolute 3.0\% on the \textsc{+Turk} set. 
The outputs of the Paraphrase and Copy Claim baselines are not Wikipedia-like, making them ineffective for augmentation. All the baseline approaches augment the false claims with a supported evidence. However, the success of our method in producing supporting evidence while trying to maintain a Wikipedia-like structure, leads to more effective augmentations.

\paragraph{Masker Analysis}

\begin{table}[t]
\centering
\begin{tabular}{c|rrr|rrr}
\toprule
$\lambda$ & \textsc{Acc} & \textsc{size} & $\Delta$ & \textsc{Prec} & \textsc{Rec} & \textsc{F1}   \\ \midrule
.5 & 5.1      & 0.0         & 5   & 0.0    & 0.0      & 0.0    \\ 
.4 & 80.0       & 26.3      & \textbf{54}  & 27.2 & 75.1   & \textbf{39.9} \\ 
.3 & 77.0       & 27.5      & 50  & 25.9 & 71.6   & 38.0   \\ 
.2 & 81.6     & 31.1      & 51  & 23.1 & 74.8   & 35.3 \\ 
\bottomrule
\end{tabular}
\caption{Results of different values of $\lambda$ for the masker with syntactic regularization. The left three columns describe the accuracy and average mask size (\% of the sentence) over the FEVER development set with the masked evidence and a neutral target label. $\Delta$ is $\textsc{Acc}-\textsc{size}$. The right three columns contain the precision, recall and F1 of the masks that we have human annotations for. For results without syntactic regularization see the appendix.}
\label{tab:mask_res}
\end{table}

To evaluate the performance of the masker model, we test its capacity to modify $\A$ and $\D$ pairs from the FEVER development set to a neutral relation. We measure the accuracy of the pretrained classifier in predicting neutral versus the percentage of masked words from the sentence. For a finer evaluation, we manually annotated 75 $\A$ and 76 $\D$ pairs with the minimal required mask for neutrality and compute the per token \textsc{F1} score of the masker against them.

The results for different values of the regularization coefficient are reported in Table \ref{tab:mask_res}. Increasing the regularization coefficient helps to minimize the mask size and to improve the precision while maintaining the classifier accuracy and the mask recall. However, setting $\lambda$ too large, can collapse the solution to no masking at all. The generation experiments use the outputs of the $\lambda=0.4$ model.
\section{Conclusion}
\label{sec:discussion}
In this paper, we introduce the task of automatic fact-guided sentence modification. Given a claim and an old sentence, we learn to rewrite it to produce the updated sentence. Our method overcomes the challenges of this conditional generation task by breaking it into two steps. First, we identify the polarizing components in the original sentence and mask them. Then, using the residual sentence and the claim, we generate a new sentence which is consistent with the claim. Applied to a Wikipedia fact update evaluation set, our method successfully generates correct Wikipedia sentences using the guiding claims. Our method can also be used for data augmentation, to alleviate the bias in fact verification datasets without any external data, reducing the relative error by 13\%.
\section{Acknowledgments}
We thank the anonymous reviewers and the MIT NLP group for their helpful discussion and comments.
This work is supported by DSO grant DSOCL18002.

\bibliography{aaai_gen}
\bibliographystyle{aaai}

\newpage
\appendix

\section{Additional Masker analysis}
\label{app:masker}

The masker model makes finding a valid mask in the space of $2^l$ options tractable. However, as mentioned in \cite{bao-etal-2018-deriving}, training an objective of the type shown in \eqref{eq:mask_gen_loss} is unstable. An alternative tractable approach is to enumerate a set of syntactic components of the evidence and score them as potential masks for neutrality. Although this approach is insufficient and might not always work, the cases where the continuous spans satisfy neutrality can help guide the masker training.

Table \ref{tab:mask_res_more} shows results for the masker model with and without syntactic regularization. The syntactic regularization helps to stabilize the performance, allowing a reasonable solution even without any additional constraint on the mask size. Without syntactic regularization, better accuracy can be achieved, but the learning is very unstable and can lead to solutions that mask the whole sentence or keep it as is.

\begin{table}[t]
\centering
\begin{tabular}{c|rrr|rrr}
\toprule
${\lambda}$ & \textsc{Acc} & \textsc{size} & $\Delta$ & \textsc{Prec} & \textsc{Rec} & \textsc{F1}   \\ \midrule
\multicolumn{7}{l}{\textit{\textbf{With} syntactic regularization:}}\\
.5 & 5.1      & 0.0         & 5.0   & 0.0    & 0.0      & 0.0    \\ 
.4 & 80.0       & 26.3      & \textbf{54}  & 27.2 & 75.1   & \textbf{39.9} \\ 
.3 & 77.0       & 27.5      & 50  & 25.9 & 71.6   & 38.0   \\ 
.2 & 81.6     & 31.1      & 51  & 23.1 & 74.8   & 35.3 \\ 
.1 & 80.5     & 34.7      & 46  & 21.9 & 77.8   & 34.2 \\ 
0   & 80.0       & 37.1      & 43  & 22.6 & 81.7   & 35.5 \\ 
\midrule
\multicolumn{7}{l}{\textit{\textbf{Without} syntactic regularization:}}\\
.5 & 5.1      & 0.0         & 5   & 0.0    & 0.0      & 0.0    \\ 
.4 & 87.8      & 25.0        & \textbf{63}   & 25.9    & 68.5      & \textbf{37.6}    \\ 
.3 & 5.1      & 0.0         & 5   & 0.0    & 0.0      & 0.0    \\ 
.2 & 90.1      & 35.0         & 55   & 22.4    & 78.7      & 34.8    \\ 
.1 & 91.2      & 48.9         & 42   & 17.0    & 85.3      & 28.4    \\ 
0   & 91.6      & 100      & -8  & 9.3 & 100   & 17.1 \\ 
\bottomrule
\end{tabular}
\caption{Results of different values of $\lambda$ for the masker with and without syntactic guidance. The left three columns describe the accuracy and average mask size (\% of the sentence) over the FEVER development set with the masked evidence and a neutral target label. The right three columns contain the precision, recall and F1 of the masks that we have human annotations for.}
\label{tab:mask_res_more}
\end{table}

\section{Example Outputs}

\begin{table*}[t]
  \small
  \centering
  
  \begin{tabular}{p{3cm}|p{11cm}}
    \toprule
      Original Text & Born in Lawton , Oklahoma and raised in Anaheim , California , Hillenburg became fascinated with the sky as a child and also developed an interest in art . \\
      Claim & Stephen Hillenburg was fascinated with the ocean as a child .\\
    \midrule
      Claim Ext. & He in Huntington , Trinidad City Tommy in the , Hillenburg developed he became the of the stage , a senior . business in the adopted in 1847 .\\
      Concat & Born in Lawton , Oklahoma and raised in Anaheim Anaheim , , Hillenburg became fascinated with the sky as a child and also developed an interest in art . \\
      M. Concat   &  Born in Lawton , Oklahoma and raised in Anaheim , California , Hillenburg became the with the United as the condition and also developed an interest in art . \\
      Ours  & Born in Lawton , Oklahoma and raised in Anaheim , California , Hillenburg became fascinated with the ocean as a child and also developed an interest in art .
  \end{tabular}
  \begin{tabular}{p{3cm}|p{11cm}}
    \toprule
      Original Text & German Startups Group considers 28 of their 42 minority stakeholdings in operationally active companies to be of particular significance to the group. \\
      Claim & It considers 23 of 43 minority stakeholdings to be significant .\\
    \midrule
      Claim Ext. & The - soon are the days the eighth capital , is the spending , , find divided active by 's the original ,\\
      Concat & German Startups Group considers 28 of their their minority stakeholdings in operationally active companies to be of particular significance to the  group . \\
      M. Concat   &  German Startups Group considers 23 of 18 minority million ` in operationally active companies to be of particular significance to the group . \\
      Ours  & German Startups Group considers 23 of 43 minority stakeholdings beginning in operationally active companies to be of particular significance to the   group .
    \end{tabular}
    \begin{tabular}{p{3cm}|p{11cm}}
    \toprule
      Original Text &  A sequel , Rio 2 , was released on April 11 , 2012 .\\
      Claim &  Rio 's sequel was released on April 11 , 2014 .\\
    \midrule
      Claim Ext. & In series , Rio is is is released on January 4 , 2014 ,\\
      Concat & A sequel , Rio Rio 2 , was released on April 11 , 2012 \\
      M. Concat   &  A sequel , Rio 2 , was released on August 11 , 2014 . \\
      Ours  & A sequel , Rio 2 , was released on April 11 , 2014 .\\
  \end{tabular}
  \begin{tabular}{p{3cm}|p{11cm}}
    \toprule
      Original Text & Albert S. Ruddy -LRB- born March 28 , 1940 -RRB- is a Canadian - born film and television producer . \\
      Claim & In 1930, Albert S. Ruddy is born.\\
    \midrule
      Claim Ext. & Albert S. S. -LRB- -LSB- Hiram 23 , 1939 -RRB- is an former actor born theoretical marketer American . .\\
      Concat & Albert S. Ruddy -LRB- born March March , , 1940 -RRB- is a Canadian - born film and television producer \\
      M. Concat   &  Albert S. Ruddy -LRB- born Hiram 12 , 1930 -RRB- is a German - American film and television producer . \\
      Ours  & Albert S. Ruddy -LRB- born December 18 , 1930 -RRB- is a Chinese - born film and television producer .\\
      \toprule
  \end{tabular}
  \caption{We compare our model outputs against different models. Each example is showing the two input sentences following the output of each model. The Concat model setting is similar to the M.\ Concat one but the original text is left unmasked. For the Claim Ext.\ model, only the claim sentence is given as input. }
  
  \label{tab:example_outputs}
\end{table*}

Examples of outputs from different models are provided in \tabref{tab:example_outputs}.
For the first 3 examples, our model produces a perfect update.
In the last example, even though our model gets the year 1930 correct, it modifies the month and nationality to made-up, incorrect values. This is a result of a too aggressive deletion by the masker.
The Claim Ext.\ model typically produces wrong and non-grammatical sentences. The Concat model doesn't capture the polarizing relation between the two inputs and mostly ignores the claim. The M.\ Concat model tends to overly generate made-up content instead of copying it from the claim.

\end{document}